\pdfoutput=1
\documentclass[conference]{IEEEtran}
\usepackage{times}

\usepackage{url}
\usepackage{verbatim}
\usepackage{graphicx}
\usepackage{color}
\usepackage{xcolor}
\usepackage{titletoc}

\usepackage[numbers]{natbib}
\usepackage{multicol}

\usepackage[
    colorlinks=true,
    linkcolor=anasblue,
    citecolor=anasblue,
    urlcolor=anasblue
]{hyperref}

\usepackage{moreverb,url}

\usepackage{amsfonts,amssymb,amsmath,bm, mathabx}
\usepackage[ruled,vlined]{algorithm2e}
\usepackage{multirow}
\usepackage{tabularx}
\usepackage{tabularray}
\UseTblrLibrary{siunitx}

\usepackage{xcolor}

\definecolor{sampler}{RGB}{209, 191, 196}
\definecolor{scalar}{RGB}{193, 197, 207}
\definecolor{vector}{RGB}{194, 204, 198}
\definecolor{logits}{RGB}{209, 198, 210}

\definecolor{redhak}{RGB}{231, 87, 83}
\definecolor{bluehak}{RGB}{5, 148, 177}
\definecolor{yellowhak}{RGB}{133, 165, 17}
\definecolor{orangehak}{RGB}{242, 142, 44}
\definecolor{greenhak}{RGB}{133, 165, 17}
\definecolor{purplehak}{RGB}{98, 84, 254}

\definecolor{anasblue}{RGB}{37, 113, 215}
\definecolor{anaspurple}{RGB}{124, 101, 181}

\usepackage{amsfonts,amssymb,amsmath,bm, mathabx}

\def\1{\bm{1}}

\def\RR{\mathbb{R}}

\def\vtheta{{\bm{\theta}}}

\def\vg{{\bm{g}}}

\def\vq{{\bm{q}}}

\def\vtheta{{\boldsymbol{\theta}}}
\def\vtau{{\boldsymbol{\tau}}}

\def\mA{{\bm{A}}}

\def\mD{{\bm{D}}}

\def\mI{{\bm{I}}}

\def\mO{{\bm{O}}}

\def\mT{{\bm{T}}}

\DeclareMathAlphabet{\mathsfit}{\encodingdefault}{\sfdefault}{m}{sl}
\SetMathAlphabet{\mathsfit}{bold}{\encodingdefault}{\sfdefault}{bx}{n}

\def\gB{{\mathcal{B}}}

\def\gN{{\mathcal{N}}}

\usepackage[]{glossaries}
\newacronym{wbc}{WBC}{Whole Body Controller}
\newacronym{hoic}{HOIC}{Hand-Object Interaction Controller}

\newacronym{dgm}{DGM}{Deep Generative Models}

\newacronym{gail}{GAIL}{Generative Adversarial Imitation Learning}
\newacronym{sqil}{SQIL}{Soft-Q Imitation Learning}

\newacronym{iflow}{iFlows}{ImitationFlows}
\newacronym{cep}{CEP}{Composable Energy Policies}
\newacronym{se3dif}{SE(3)-DiF}{SE(3)-DiffusionFields} 

\newacronym{nlp}{NLP}{Natural Language Processing}
\newacronym{cv}{CV}{Computer Vision}

\newacronym{gp}{GP}{Gaussian Process}
\newacronym{gan}{GAN}{Generative Adversarial Networks}
\newacronym{vae}{VAE}{Variational Autoencoders}
\newacronym{cvae}{cVAE}{Conditional Variational Autoencoders}
\newacronym{ebm}{EBM}{Energy Based Models}
\newacronym{sbm}{SBM}{Score based Models}
\newacronym{nf}{NFlow}{Normalizing Flows}

\newacronym{dm}{DM}{Diffusion Models}
\newacronym{ddpm}{DDPM}{Denoising Diffusion Probabilistic Models}
\newacronym{ncsn}{NCSN}{Noise Conditioned Score Network}
\newacronym{smld}{SMLD}{Score Matching with Langevin Dynamics}

\newacronym{map}{MAP}{Maximum a Posteriori}

\newacronym{mdm}{MDM}{Mixture Density Models}
\newacronym{gmm}{GMM}{Gaussian Mixture Models}

\newacronym{nerf}{NeRF}{Neural Radiance Fields}

\newacronym{bc}{BC}{Behavioural Cloning}
\newacronym{il}{IL}{Imitation Learning}
\newacronym{irl}{IRL}{Inverse Reinforcement Learning}
\newacronym{ioc}{IOC}{Inverse Optimal Control}
\newacronym{lfd}{LfD}{Learning from Demonstration}
\newacronym{em}{EM}{Expectation Maximization}
\newacronym{promp}{ProMP}{Probabilistic Movement Primitives}
\newacronym{dmp}{DMP}{Dynamic Movement Primitives}
\newacronym{seds}{SEDS}{Stable Estimator of Dynamical Systems}
\newacronym{gmr}{GMR}{Gaussian Mixture Regressor}
\newacronym{gpr}{GPR}{Gaussian Process Regressor}
\newacronym{lwr}{LWR}{Locally Weighted Regressor}
\newacronym{kmp}{KMP}{Kernelized Movement Primitives}
\newacronym{clf}{CLF}{Control Lyapunov Function}
\newacronym{wsaqf}{WSAQF}{Weighted Sum of Asymmetric Quadratic Function)}
\newacronym{nilc}{NILC}{Neurally Imprinted Lyapunov
Candidate}
\newacronym{clfdm}{CLF-DM}{Control Lyapunov Function-based Dynamic Movements}
\newacronym{cnmp}{CNMP}{Conditional Neural Movement Primitives}
\newacronym{tpgmm}{TP-GMM}{Task Parameterized GMM}

\newacronym{mp}{MP}{Movement Primitive}
\newacronym{mpflows}{MPFlows}{Movement Primitive Flows}

\newacronym{gcl}{GCL}{Guided Cost Learning}

\newacronym{mle}{MLE}{Maximun Likelihood Estimation}
\newacronym{sde}{SDE}{Stochastic Differential Equation}
\newacronym{ode}{ODE}{Ordinary Differential Equation}

\newacronym{probs}{ProbS}{Probabilistic Segmentation}
\newacronym{crf}{CRF}{Conditional Random Fields}
\newacronym{ppca}{PPCA}{Probabilistic Principal Component Analysis}
\newacronym{gmcc}{GMCC}{Generalized Multiple Correlation Coeficcient}
\newacronym{hri}{HRI}{Human-Robot Interaction}
\newacronym{ip}{IP}{Interaction Primitives}
\newacronym{hmm}{HMM}{Hidden Markov Model}
\newacronym{cac}{CAC}{Canonical Correlation Coefficient}
\newacronym{rv}{$R_v$}{$R_v$ Coefficient}
\newacronym{dcor}{dCor}{Distance Correlation}
\newacronym{dtw}{DTW}{Dynamic Time Warping}
\newacronym{edr}{EDR}{Edit Distance With Real Penalty}
\newacronym{twed}{TWED}{Time Warp Edit Distance}
\newacronym{r2}{$R^2$}{Coefficient of Determination}
\newacronym{sqp}{SQP}{Successive Quadratic Programming}
\newacronym{rkhs}{RKHS}{Reproducing Kernel Hilbert Space}
\newacronym{icnn}{ICNN}{Input-Convex Neural Network}

\newacronym{pca}{PCA}{Principal Component Analysis}

\newacronym{maf}{MAF}{Masked Autoregressive Flow}
\newacronym{iaf}{IAF}{Inverse Autoregressive Flow}
\newacronym{node}{N-ODE}{Neural ODE}
\newacronym{nsflow}{NSF}{Neural Spline Flows}
\newacronym{cnf}{CNF}{Conditional Normalizing Flows}
\newacronym{ffjord}{FFJORD}{Free-form Jacobian of Reversible Dynamics}

\newacronym{inn}{INN}{Invertible Neural Networks}
\newacronym{mcmc}{MCMC}{Markov Chain Monte Carlo}
\newacronym{ld}{LD}{Langevin Dynamics}

\newacronym{cd}{CD}{Contrastive Divergence}
\newacronym{nce}{NCE}{Noise Contrastive Estimation}
\newacronym{ce}{CE}{Cross-Entropy}
\newacronym{dsm}{DSM}{Denoising Score Matching}

\newacronym{ik}{IK}{Inverse Kinematics}

\newacronym{sdf}{SDF}{Signed Distance Field}
\newacronym{deepsdf}{DeepSDF}{Deep Signed Distace Field}

\newacronym{emd}{EMD}{Earth Mover Distance}

\newacronym{mlp}{MLP}{Multi Layer Perceptron}
\newacronym{rrt}{RRT}{Rapidly-exploring Random Trees}

\newacronym{relu}{ReLU}{Rectified Linear Unit}

\newacronym{vn}{VN}{Vector Neuron}

\newacronym{icp}{ICP}{Iterative Closest Point}

\newacronym{rmp}{RMP}{Riemannian Motion Policies}
\newacronym{mpc}{MPC}{Model Predictive Control}

\newacronym{svi}{SVI}{Structured Variational Inference}
\newacronym{vi}{VI}{Variational Inference}
\newacronym{hrl}{HRL}{Hierarchical Reinforcement Learning}
\newacronym{apf}{APF}{Artificial Potential Fields}

\newacronym{reps}{REPS}{Relative Entropy Policy Search}
\newacronym{rl}{RL}{Reinforcement Learning}

\newacronym{lmdp}{LMDP}{linearly-solvable Markov Decision Processes}
\newacronym{dwa}{DWA}{Dynamic Window Approach}
\newacronym{svf}{SVF}{Stable Vector Fields}

\newacronym{mse}{MSE}{Mean Squared Error}

\newacronym{stomp}{STOMP}{Stochastic Trajectory Optimization for Motion Planning}
\newacronym{gpmp}{GPMP}{Gaussian Process Motion Planning}
\newacronym{chomp}{CHOMP}{Covariant Hamiltonian Optimization for Motion Planning}

\newacronym{llm}{LLM}{Large Language Models}
\newacronym{vlm}{VLM}{Vision-Language Model}

\usepackage{cleveref}

\usepackage{xspace}
\usepackage{capt-of}
\newcommand{\ourMethod}{\mbox{Dex-S0}\xspace}

\let\oldcite\cite
\renewcommand{\cite}[1]{\textbf{\oldcite{#1}}}
\let\oldcitep\citep
\renewcommand{\citep}[1]{\textbf{\oldcitep{#1}}}

\makeatletter
\let\@oldmaketitle\@maketitle%
\renewcommand{\@maketitle}{\@oldmaketitle%
    \vspace{5.0mm}
    \setcounter{figure}{0}
    \centering
    \includegraphics[width=.99\linewidth]{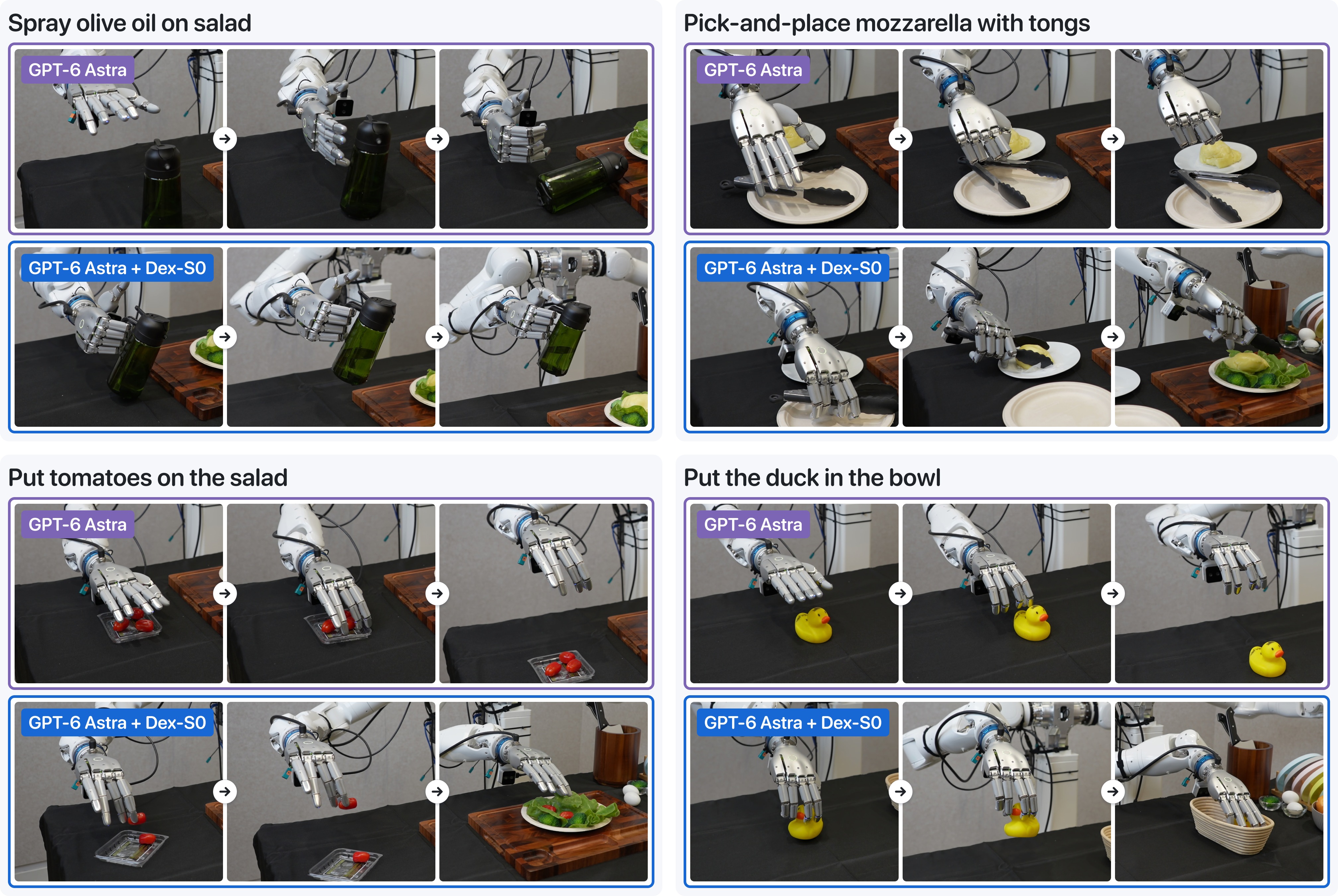}
    \captionof{figure}{\textbf{\ourMethod turns an agent's coarse commands into precise dexterous actions.} Each box is a real task, with time running from left to right. GPT-6 Astra executed directly (top) struggles, while the same agent through \ourMethod (bottom) completes the task.}
    \label{fig:main}
    \vspace{-.35cm}
}
\makeatother

\begin{document}

\title{\fontsize{19.2}{23}\selectfont \textbf{\ourMethod}: A Hand-Object Interaction Controller \\
to Map Coarse Behaviors into Fine-Grained Actions}

\makeatletter\def\@IEEEauthorblockAtopspace{0.75ex}\makeatother
\author{\IEEEauthorblockN{\mbox{Anas Houssaini}\hspace{1.4em}\mbox{Fran\c{c}ois Hogan}\hspace{1.4em}\mbox{Kenneth Shaw}\hspace{1.4em}\mbox{Mustafa Mukadam}\\\mbox{Nur Muhammad Mahi Shafiullah}\hspace{1.4em}\mbox{Homanga Bharadhwaj}\hspace{1.4em}\mbox{Jitendra Malik}\hspace{1.4em}\mbox{Julen Urain}}

\IEEEauthorblockA{Amazon FAR}}

\maketitle

\begin{abstract}
While agentic models excel at generating coarse robotic behaviors, their direct application to dexterous manipulation is bottlenecked by a lack of precision and physical grounding.
These high-level models successfully plan macro-intentions, such as reaching toward a target, but struggle to resolve the fine-grained interactions required for task execution, like coordinating individual fingers to actuate a spray bottle.
We present \ourMethod, a local hand-object interaction controller designed to translate coarse behavioral priors into precise dexterous actions.
Formulated as a downstream local policy, \ourMethod takes a semantic skill label and a coarse motion trajectory from a generalist policy as input.
By grounding these commands with proprioceptive feedback and local wrist-depth observations, \ourMethod dynamically adapts the coarse trajectory into the multi-finger action sequence required to solve the task.
We evaluate \ourMethod on dexterous simulated and real-world tasks.
Project page: \href{https://dexs0.com}{dexs0.com}.

\end{abstract}
\IEEEpeerreviewmaketitle

\begin{figure*}[t]
    \centering
    \includegraphics[width=1.0\linewidth]{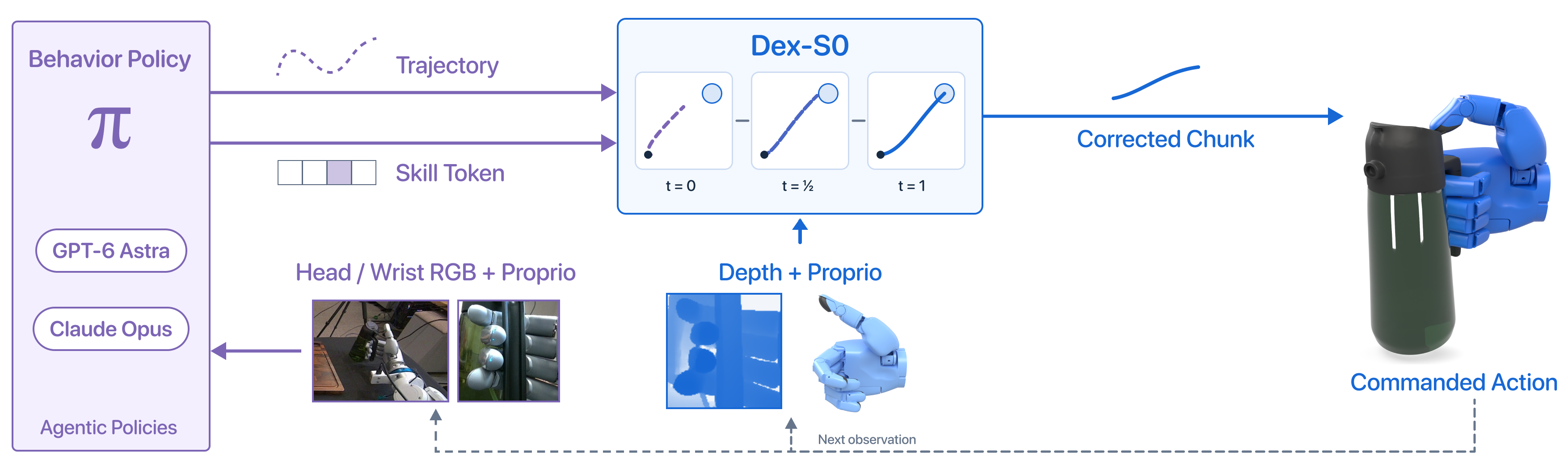}
    \caption{\textbf{\ourMethod architecture.} The behavior policy, GPT-6 Astra or Claude Opus 5.5, sends a trajectory of the hand for 2 seconds and a skill label from a fixed set (reach, grasp, press, release, \dots). 
    \ourMethod starts denoising from the reference trajectory for the wrist and the fingers ($t{=}0$), conditioned on the skill token, wrist depth and finger proprioception, and denoise the trajectory to the closest feasible actions ($t{=}1$). The robot executes the first 0.5 seconds of the corrected chunk, and \ourMethod runs again on the new observation.}
    \label{fig:hoir_architecture}
    \vspace{-5mm}
\end{figure*}

\section{Introduction}
\label{sec:intro}
Dexterous manipulation fails on small errors near contact.
A hand can move correctly for most of a task and still lose the object because a finger approaches from the wrong side or makes contact a few centimeters off target.
These errors matter more as tasks involve tool use or interacting with articulated objects.
The problem is timely because generalist models, such as \textbf{agentic models}, can now reason over long-horizon tasks~\citep{ahn2022saycan, liang2022codepolicies, huang2022innermonologue} and produce plausible 3D hand motions~\citep{huang2023voxposer}, but they \textbf{fail on the last mile} (\Cref{fig:main}).
Solving complex dexterous manipulation tasks requires both coarse reasoning about the behavior and dexterous action to grasp and use tools.
This leads to the question that motivates this work.

\textit{How can we use the coarse reasoning of generalist agentic models and still solve dexterous manipulation tasks?}

We introduce \textbf{Dexterity-System 0 (\ourMethod)}, a policy that maps coarse behavioral inputs to fine-grained dexterous actions (\Cref{fig:hoir_architecture}).
\ourMethod receives two types of inputs.
\\
(\textbf{a}) \textbf{Coarse intent commands} generated by a high-level behavioral policy, such as an agentic model.
\\
(\textbf{b}) \textbf{Environment observations} focused on informing about the local hand-object interactions in the scene.
\\
\ourMethod generates actions that follow the high-level intent while adapting to the local scene observations.
We study two questions. (\textbf{1}) Can \ourMethod take coarse, imprecise commands and correct them into successful dexterous actions? (\textbf{2}) Does this let general-purpose agentic models solve dexterous tasks they rarely solve alone?

With \ourMethod, we show that agentic models can successfully perform contact-rich, dexterous manipulation and tool-use in the real world (\Cref{fig:main}).
Two design choices make it work.
\\
\textbf{Intent as a skill token and a coarse trajectory.} A coarse motion trajectory alone is not sufficient to convey intent. Geometrically similar motions can require opposite finger behaviors (e.g., pressing vs.\ releasing a trigger). Adding a semantic skill token disambiguates the desired interaction, which matters for contact-rich tasks such as tool use.
\\
\textbf{Cropped wrist depth.} Proprioception informs \ourMethod where the fingers are, but not where the object is. A wrist-mounted depth camera, cropped to the local interaction region, gives \ourMethod the relative hand-object geometry it needs to correct the approach \emph{before} contact happens (\Cref{fig:grasp_token}). Because the observation is local and wrist-centered, it generalizes across scenes without depending on external camera placement.
In our ablations, removing wrist depth or the skill token lowers the average success from 93\% to at most 13\%.

Architecturally, \ourMethod is a diffusion transformer that uses the behavior policy's trajectory to seed the denoising process, analogous to img2img generation~\citep{meng2021sdedit, couairon2022diffedit, yoneda2023noise, park2026demodiffusion}. \ourMethod trains only on robot play data interacting with the object or tool and requires no paired data from the upstream policy, making it agnostic to the source of behavior.

We evaluate \ourMethod on four simulated DexSuite~\citep{dexsuite} tasks and five real-world tasks. With GPT-6 Astra~\citep{openai2026gpt6astra} and Claude Opus 5.5~\citep{anthropic2026opus55}, \ourMethod raises success from 48\% to 88\% on pick-and-place tasks and from 3\% to 63\% on tasks that require high precision and tool use such as  spraying and toasting. When the agents learn from their own attempts, GPT-6 Astra learns every task through \ourMethod, but never learns spraying or toasting without \ourMethod routing the trajectory into grounded actions.

\section{Related Work}
\label{sec:related_work}

\textbf{From Coarse Commands to Precise Actions.}
The challenge of translating coarse high-level commands into precise low-level execution is well studied in locomotion, where \glspl{wbc} map kinematic references or contact goals into physically feasible joint commands~\citep{he2024learning, wococo, fu2023deep, halo2026}.
In manipulation, \glspl{hoic} capture the same idea, a local controller dedicated to the hand-object interface.
Residual policies learn such a correction on top of a base controller~\citep{silver2018residual, johannink2019residual}.
DexterityGen~\citep{yin2025dexteritygen} introduced the concept of a foundational dexterity layer, a controller trained in simulation that adapts coarse motion commands into dexterous actions using proprioceptive feedback.
However, coarse motion commands alone may not capture how the environment should change as a result of the interaction.
Explicitly representing this \emph{intent} helps the controller disambiguate the desired outcome.
PianoMime~\citep{qian2024pianomime} augments the desired motion commands with a sequence of desired piano keys, making the musical intent explicit and enabling the play of arbitrary songs.
Track2Act~\citep{bharadhwaj2024track2act} and Im2Flow2Act~\citep{xu2024flow} use point tracks and object flow extracted from video as a cross-embodiment action interface, while
LUCID~\citep{gupta2026lucid} defines intent through 3D object flow.
\ourMethod pairs a coarse motion trajectory with a \emph{semantic} skill token (e.g., grasp, press, release) that disambiguates intent when geometry alone is insufficient, for instance to distinguish pressing from releasing a trigger.
Skill tokens are a natural interface for agentic models, which easily output semantic action labels alongside coarse trajectories.

\textbf{Agentic Models for Manipulation.}
During the last year, we have found how agentic models have been established as an alternative option to Vision-Language-Action models~\citep{brohan2023rt2, kim2024openvla, black2024pi0, intelligence2025pi05, geminirobotics2025} for robot action generation.
On the RoboDojo benchmark~\citep{chen2026robodojo}, pure agentic systems rank among the top performers~\citep{zhang2026gpt6astra}. PyRUA-Lean~\citep{si2026pyrualean}, HexaShell~\citep{gao2026hexashell}, and EmbodiedEvo~\citep{li2026embodiedevo} achieve up to 70\% average success by combining \glspl{vlm} with code generation and scene reasoning.
However, when these agents are evaluated on dexterous, contact-rich tasks, they consistently underperform (see \Cref{fig:main}).
\ourMethod is designed to close this gap. It sits downstream of any agentic model and converts its coarse outputs into precise contact-rich actions.

\section{\ourMethod: A Hand-Object Interaction Controller}
\label{sec:hoir}

\begin{figure}[t]
    \centering
    \includegraphics[width=\linewidth]{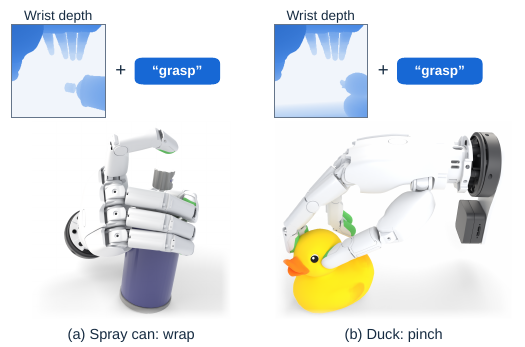}
    \caption{\textbf{One skill token, two grasps.} Both hands receive the token ``grasp''. From the wrist depth at the approach, \ourMethod wraps the spray can (a) and pinches the duck by the head (b).}
    \label{fig:grasp_token}
    \vspace{-2mm}
\end{figure}

\begin{figure*}[t]
    \centering
    \includegraphics[width=\textwidth]{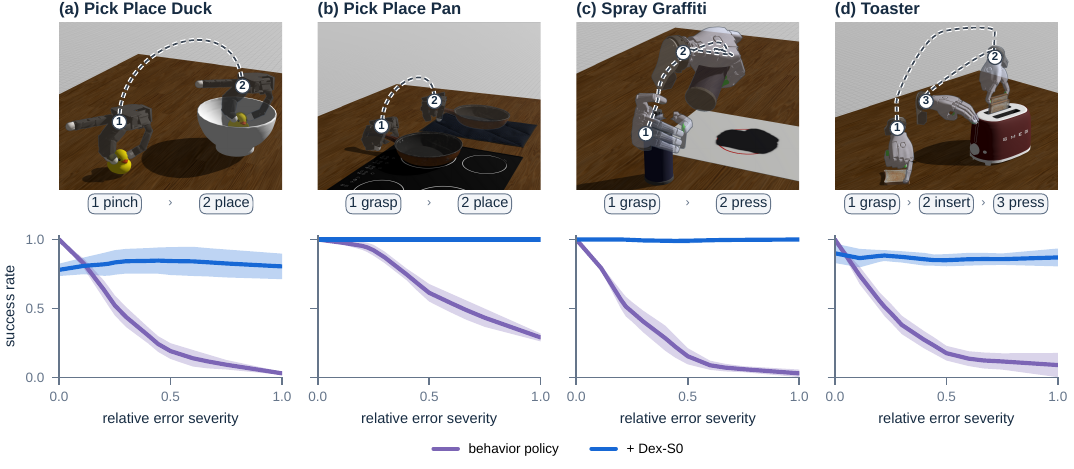}
    \caption{\textbf{With \ourMethod, the behavior policy does not need to be precise.} (Top) The four tasks and when the skills are activated. (Bottom) We corrupt the behavior policy's commands with growing errors to test whether \ourMethod recovers the interaction from a bad intent. Executed directly (purple), the behavior policy fails once its errors exceed what the contact tolerates. Through \ourMethod (blue), it keeps succeeding. Success is averaged over the four error types (\Cref{tab:noise}), with 10 episodes per level and bands of one standard deviation over five seeds.}
    \label{fig:behavior_degradation}
    \vspace{-2mm}
\end{figure*}

A Hand-Object Interaction Controller (\Gls{hoic}) is a goal-conditioned policy $\pi_{\vtheta}(\mA|\mO, \vg)$ that generates a sequence of dexterous actions $\mA$, conditioned on a set of observations $\mO$ and some high-level goal/intention $\vg$.
A \gls{hoic} is \textbf{local}.
In contrast with a general policy which must process broad contextual information to understand a scene and plan task-solving behaviors, a \gls{hoic} strictly manages localized hand-object dynamics. It relies on a higher-level policy $\pi_{\gB}$ to dictate the overarching behavioral strategy, and its goal is to follow the behavioral intent while correctly solving the hand-object interactions.

We first describe the desiderata of a good \gls{hoic} and then design \ourMethod to satisfy them.

\subsection{What Makes a Good Hand-Object Interaction Controller?}
\label{sec:good_hoic}
We found two desirable properties for a good \gls{hoic}, locality and a healthy balance between following the high-level intent and correcting it.
\\
\textbf{Locality.}
The intuitive desirable property of a strong \gls{hoic} is to be local, focused on the hand-object interactions while a higher-level behavior policy focuses on the general task.
However, locality is a broad term and the selection of the observations and the goals depends on how precise the high-level behavior policy is.
Given a precise high-level behavior policy that informs on the exact finger placement, a \gls{hoic} can rely on purely local observations such as proprioception, force and touch, and only correct the force interactions. A coarser behavior policy might require a more capable \gls{hoic}, with access to local depth or RGB signals to solve the task.
\\
\textbf{Balance between follow and correct.} 
A strong \gls{hoic} balances how much to follow the higher-level intent and how much to correct it.
As with locality, this balance depends on how precise the higher-level behavior policy is.
For example, a behavior policy with a precision error of a few centimeters might be good for general motion intent (reach to the drawer, pick this object), but closely imitating it for precise insertion or pressing a button might lead to failure. 
Thus, building a useful \gls{hoic} requires a clear understanding of the limits of the behavior policy to adapt accordingly.

Both properties, and their relation to the quality of the behavior policy, are intertwined. Consider, for example, the task of pressing a spray bottle. If the upstream plan places the finger on the trigger, a good \gls{hoic} only needs local force modulation to actuate it, but if the plan lands the finger a few centimeters off on the bottle neck, no amount of force correction recovers, and the \gls{hoic} must re-localize from wrist depth and shift the hand before committing.

\subsection{Building \ourMethod}
The design of our \gls{hoic} cannot be done independently of the behavior policy, as the accuracy of the higher-level policy directly influences the design choices for interacting with it.
\ourMethod is designed to interact with a particular type of behavior policy, agentic models.
We make three assumptions on the behavior policies.
\begin{itemize}
    \item \textbf{Wrist pose targets} accurate to within a few centimeters.
    \item \textbf{Coarse finger placements}, capturing general intentions such as close/open.
    \item \textbf{A skill token}, describing the primitive desired to be executed (e.g., reach, grasp, press). 
\end{itemize}

The skill token serves two purposes. First, it adds information on the desired intent that a coarse motion might fail to convey. Second, it informs on the intent beyond the robot motion. A similar robot motion might have different outcomes on the object it interacts with. The skill token helps disambiguate what the behavior policy aims to do with respect to the object it interacts with.

\subsubsection{Observations, Actions, \& Goals}
We select the observations and goals based on the expected precision of our behavior policies. Both observations and actions are local, centered around the current wrist frame.
\\
\textbf{Observations.} \ourMethod is conditioned only on local observations. In particular, we consider finger proprioception $\vq_c$ and cropped wrist depth $\mD \in \RR^{W\times H}$. 
While finger proprioception informs about the robot state, cropped wrist depth allows localizing the robot hand relative to its local environment, as in vision-based dexterous grasping~\citep{lum2024dextrah, singh2024dextrah, weng2024dexdiffuser}, and correcting the behavior policy's intent, enhancing the hand-object interaction.
We do not include the wrist pose as an observation, since all actions and observations are relative to it.
\\
\textbf{Actions.} \ourMethod predicts a trajectory of commanded actions $\mA = (\vtau_{\mT}, \vtau_{\vq})$ for both the robot's wrist pose $\mT$ and finger joints $\vq$.
The wrist commands $\vtau_\mT = \mT_{1:K}$ are a trajectory of $K$ poses $\mT \in SE(3)$ centered around the current wrist frame. The finger configurations $\vtau_{\vq}=\vq_{1:K}$ are a trajectory of $K$ joint configurations $\vq$ in absolute positions.
During deployment, wrist commands are mapped to the arm joints through inverse kinematics \citep{pink}.
\\
\textbf{Intent/Goal.} The behavior policy informs about the desired intent through two variables, a trajectory of desired configurations and a semantic skill token.
The trajectory of desired configurations is composed of desired wrist poses and finger poses.
The semantic skill token is a one-hot encoding vector informing on the primitive skill (e.g., reach, grasp, press) the behavior policy intends to execute.
While the trajectory informs geometrically on the desired motion, the semantic skill token adds context on the desired behavior. 
The behavior policy does not inform on object behaviors and can be coarse in its motion commands, particularly for the fingers. The skill token helps induce the proper behavior despite this missing geometric information.
\\
\textbf{The token says what, the depth says how.} A skill token says nothing about the object. The behavior policy sends the same ``grasp'' for a spray can and for a rubber duck, and \ourMethod turns it into a functional grasp~\citep{agarwal2023dexterous, huang2024fungraspfunctionalgraspingdiverse} of each, from what its wrist camera sees (\Cref{fig:grasp_token}). The behavior policy only has to name the skill and bring the hand close.

\subsubsection{Training Data}

Since \ourMethod is local (its observations are limited to local depth and proprioception), the training data does not require long-horizon tasks, only local hand-object interactions, enough to correct the behavior policy's commands.
We first collect some demonstrations of the different skills we aim to teach to our \ourMethod policy, with around 50 demonstrations per skill.
We add two post-processing steps, demonstration labeling and synthetic augmentation, to prepare the data for \ourMethod.
\\
\textbf{Demonstration labeling.} We label the collected training data with skill tokens. 
For each demonstration, we call a Vision-Language Model (VLM) to segment and label our collected trajectories with the possible skills \ourMethod can apply.
\begin{figure*}[t]
    \centering
    \includegraphics[width=\textwidth]{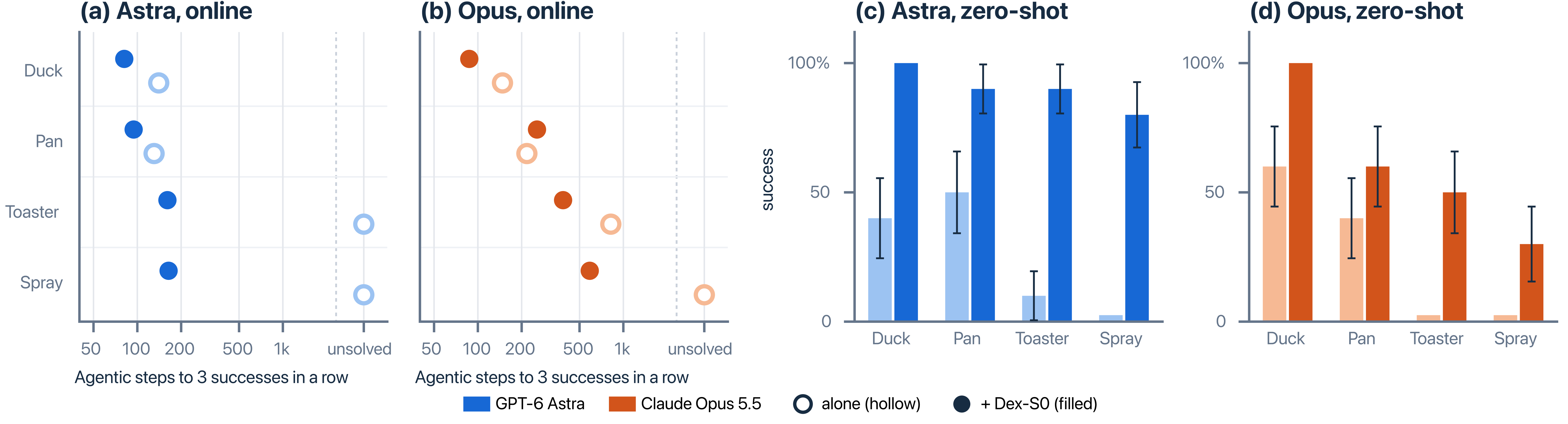}
    
    \caption{\textbf{\ourMethod with agentic policies.} (a, b) Agentic steps an agent needs to solve a task three times in a row, summed over its episodes and averaged over the runs that get there. One episode takes between 11 and 80 agentic steps, at most 35 for the pan, 60 for the duck and the spray and 80 for the toaster. Unsolved means most of the three runs never do within 20 episodes. (c, d) Zero-shot success over 10 episodes, with the standard error. Filled dots and dark bars show the agent through \ourMethod, hollow dots and light bars the agent alone.}
    \label{fig:agentic}
    \vspace{-0.3cm}
\end{figure*}
\\
\noindent\textbf{Synthetic Data Augmentation.} We found that with a small training dataset, \ourMethod can bias the commanded motion toward the few motions it has seen.
This is particularly evident in some parts of the task. 
When the robot is hovering around or when it has already picked an object and is moving with it, \ourMethod needs a large coverage of behaviors. The behavior policy might want to move in any direction, and \ourMethod should follow it.
To achieve this, we need to guarantee that our training dataset has enough coverage of those motions.
On the other hand, in the states where a dexterous hand-object interaction is about to happen (grasping, pressing a button, using a tool), the distribution of behaviors should be narrower, projecting an imprecise behavior policy onto the few modes in \ourMethod's repertoire.
In these situations dexterity is required, and we should rely on \ourMethod's corrections over the behavior policy's intent.
Thus, we apply targeted synthetic augmentations per skill, to grow the coverage of \ourMethod's behaviors where needed.

\subsection{Policy Architecture}

We build \ourMethod on DiT policy~\citep{dasari2024ditpi}, a Diffusion Policy~\citep{chi2025diffusion} with a Diffusion Transformer (DiT) backbone~\citep{peebles2023scalable}, and adapt it for dexterous manipulation.
\\
\textbf{Coarse \& Fine-grain conditioning.}
The behavior intent is given through both geometric motion trajectories and semantic skill tokens. 
We divide the goal conditioning between the coarse skill semantic label and the fine-grain motion intent.
At inference, \ourMethod does not start from pure noise. Similarly to img2img diffusion models~\citep{meng2021sdedit, couairon2022diffedit} or diffusion-based shared autonomy~\citep{yoneda2023noise, park2026demodiffusion}, the wrist actions of the initial sample $\mA_0$ are the behavior policy's motion trajectory. The finger actions are drawn from $\gN(\mathbf{0}, \mI)$. The semantic skill token conditions the denoising process through AdaLN~\citep{peebles2023scalable}. Intuitively, it tells \ourMethod what the fingers must do, since they start from noise, and which corrections the wrist needs, even though the wrist starts from the behavior policy's trajectory.
With this conditioning, training needs no paired data between the behavior policy's motion and the actions of \ourMethod.
We only need to label the demonstrations with their skills.

\section{Experimental Evaluation}
\label{sec:exps}

We evaluate \ourMethod on three questions.
(\textbf{1}) Can \ourMethod correct coarse commands whose intent is right but whose execution is imprecise?
(\textbf{2}) Does \ourMethod let general-purpose agentic models solve dexterous tasks they rarely solve alone, in simulation and on a real robot?
(\textbf{3}) Which components of \ourMethod matter?

\subsection{Performance under Imperfect Behavior Policies}
\label{sec:imperfect_policies}

\begin{figure}[ht]
\centering
\includegraphics[width=\textwidth]{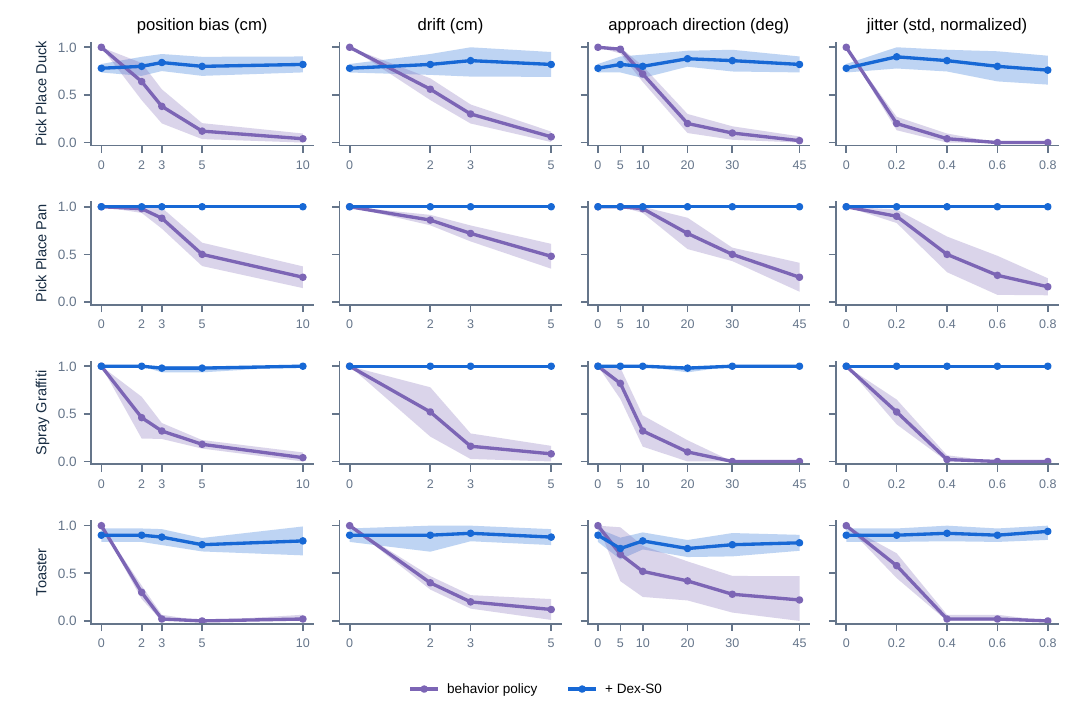}
\caption{Task success for every error type and magnitude, the behavior policy executed directly (purple) and routed through \ourMethod (blue), 10 episodes per point; 0 is no error. Each curve is a mean with a shaded band of one standard deviation, over five training seeds for \ourMethod and over five sets of evaluation episodes for the behavior policy.}
\label{fig:per_error_type}
\end{figure}

\ourMethod recovers behavior policies that have the right intent but an imprecise execution. We take an oracle behavior policy, corrupt its commands with growing errors, and execute it either directly or through \ourMethod (\Cref{fig:behavior_degradation}).
\\
\textbf{Tasks.}
We use four DexSuite~\citep{dexsuite} tasks, each built around a contact-critical interaction (\Cref{fig:behavior_degradation}, top).
In \textbf{Pick-and-place duck}, the robot pinches a rubber duck by the head and places it in a bowl.
In \textbf{Pick-and-place pan}, it grasps a pan by the handle and moves it from a hot plate to a kitchen mat.
In \textbf{Spray graffiti}, it picks up a spray can, aims it at a sheet of paper and sprays until the paint covers 60\% of the marked area.
In \textbf{Toaster}, it pinches a slice of toast, inserts it into the slot and pushes down the lever.
\\
\textbf{Behavior policy and errors.}
The oracle is a scripted controller with access to the true state of the environment, and it solves nearly every episode. \ourMethod is trained on its successful rollouts and only sees proprioception and wrist depth.
While the hand approaches a contact, we corrupt the oracle's commands with a position bias, a drift, a tilt of the approach direction or jitter, at increasing magnitudes (Appendix~\ref{app:noise}).
\\
\textbf{Results.}
Executed directly, the behavior policy fails quickly as its commands become imprecise. A few centimeters or degrees of error near contact are often enough (\Cref{fig:behavior_degradation}, bottom).
Routed through \ourMethod, the same commands keep succeeding over the whole range of errors, in all four tasks and for every error type (\Cref{tab:noise_results} and \Cref{fig:per_error_type}).
The behavior policy only has to bring the hand to the object and choose the skill. \ourMethod solves the interaction.

\begin{figure}[ht]
    \centering
    \includegraphics[width=\columnwidth]{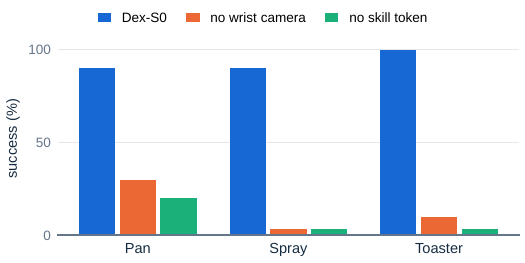}
    \vspace{-2mm}
    \caption{\textbf{Ablations.} Zero-shot success of GPT-6 Astra through \ourMethod and through variants without wrist depth or without the skill token, over 10 episodes per task. For the spray and the toaster, all three are trained together on the pan, spray and toaster data.}
    \label{fig:ablations}
    \vspace{-3mm}
\end{figure}

\begin{figure*}[t]
    \centering
    \includegraphics[width=\textwidth]{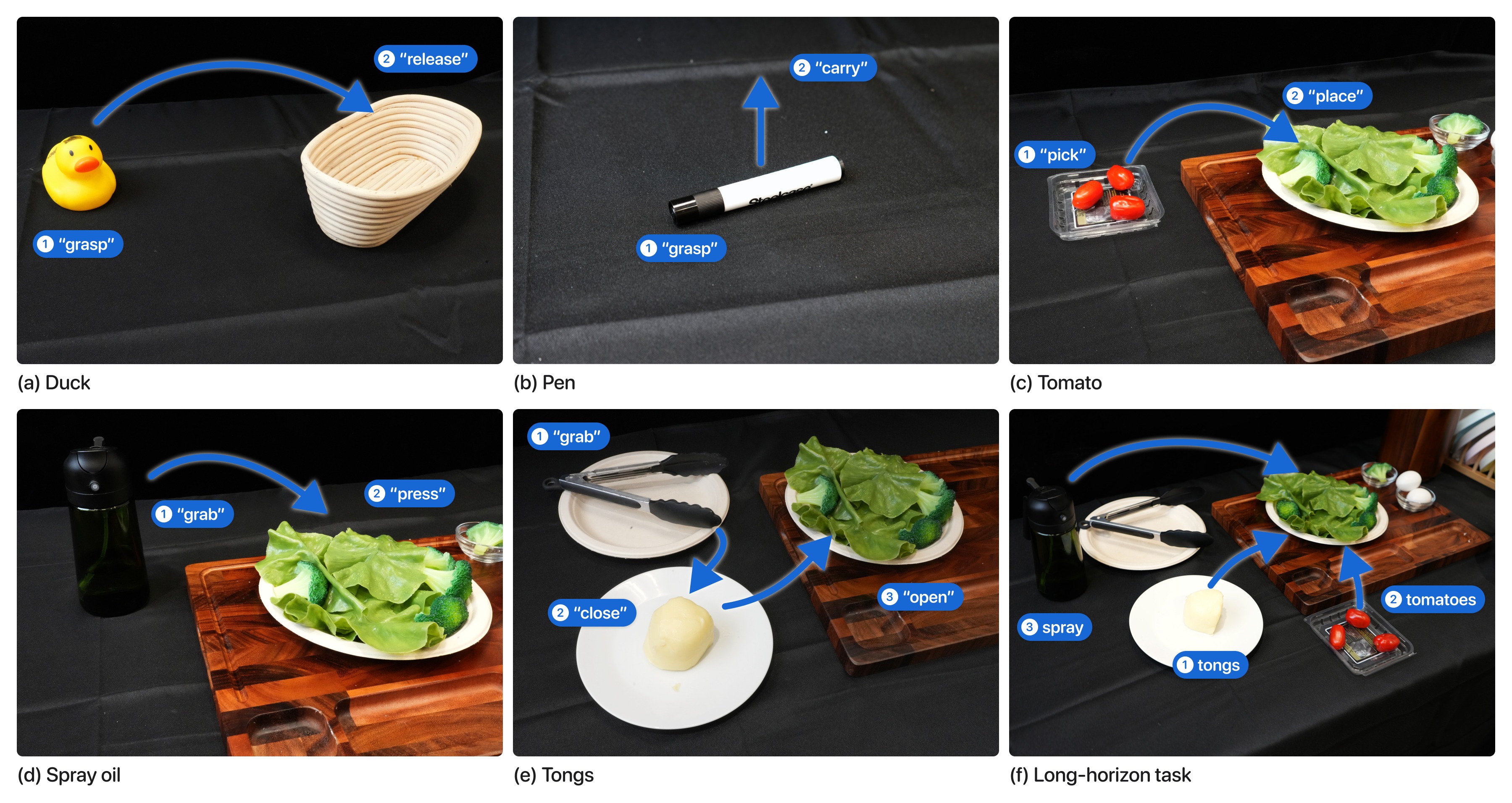}
    \vspace{-2mm}
    \caption{\textbf{Real-world tasks.} (a--e) The five real-world tasks. The arrows show the motion of each task and the numbered labels the order of the skills \ourMethod executes. (f) A long-horizon task that chains the tongs, tomato and spray tasks to make a salad, shown qualitatively.}
    \label{fig:real_tasks}
    \vspace{-3mm}
\end{figure*}

\subsection{\ourMethod with Agentic Policies}
\label{sec:agentic_policies}

With \ourMethod, general-purpose agentic models solve precise dexterous tasks they almost never complete alone.
The behavior policy is now GPT-6 Astra or Claude Opus 5.5. It sees the robot's cameras, its proprioception and a short task description with advice, and decides by itself how to solve the task. It acts in agentic steps. At each step it answers with the skill to use and where the hand should be after 1 and 2 seconds. We interpolate these two poses into a 2 second trajectory, limited to the robot's speed, and execute its first 0.5 seconds before the next step. The trajectory runs either directly on the robot (\textit{agent alone}), where the agent also sets the fingers, or through the same \ourMethod as in \Cref{sec:imperfect_policies} (\textit{agent + \ourMethod}), on the same four tasks.
An episode ends when the agent says the task is done or when it reaches the step limit of the task. In \textit{zero-shot execution}, every episode starts from an empty context. In \textit{iterative in-context learning}, the agent keeps its context across episodes and is told after each one whether it succeeded.
\\
\textbf{Zero-Shot Execution.}
On the two pick-and-place tasks, the agents alone already succeed in 40--60\% of the episodes, but they usually need many retries. \ourMethod adds 40 percentage points on average.
On the precise tasks the agents alone almost never succeed, because a small finger misplacement is enough to miss the spray nozzle or the toaster slot. Through \ourMethod, GPT-6 Astra reaches 80\% on the spray and 90\% on the toaster, and Claude Opus 5.5 reaches 30\% and 50\% (\Cref{fig:agentic}, c, d).
\\
\textbf{Iterative In-Context Learning.}
Each run is one context of up to 20 episodes, and a task counts as learned after three successful episodes in a row. Every agent and condition has three independent runs. We count the agentic steps over all episodes until the task is learned (\Cref{fig:agentic}, a, b).
Through \ourMethod, GPT-6 Astra learns all four tasks in every run, in 80 to 165 agentic steps on average. Alone, it learns the two pick-and-place tasks in about 130 to 140 steps but never learns the toaster or the spray.
Claude Opus 5.5 learns more slowly. Through \ourMethod, it learns the toaster in less than half the steps it needs alone, and it learns the spray in two of its three runs, against one alone.

\begin{figure}[ht]
    \centering
    \includegraphics[width=\columnwidth]{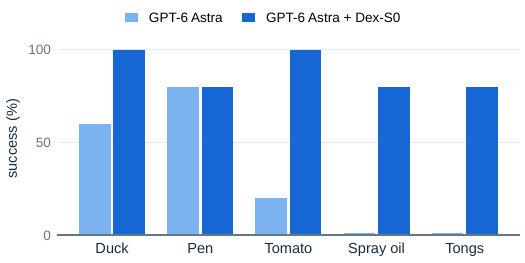}
    \vspace{-2mm}
    \caption{\textbf{Real-world results.} Task success of GPT-6 Astra executed directly (light blue) and through \ourMethod (blue), over 5 trials per task.}
    \label{fig:real_results}
    \vspace{-3mm}
\end{figure}

\subsection{Ablation Studies}
\label{sec:ablations}

We train two variants of \ourMethod, each without one component. The first has no wrist depth and keeps only proprioception, and the second has no skill token. Both are evaluated zero-shot with GPT-6 Astra, as in \Cref{sec:agentic_policies} (\Cref{fig:ablations}).

Wrist depth and the skill token both matter. Without wrist depth, the average success drops from 93\% to 13\%, and the object leaves the table in only 9 of 30 episodes. Proprioception alone does not tell \ourMethod where the object is, and the grasp fails.
Without the skill token, the agent sets the fingers itself, as when it acts alone, and the average success drops to 7\%. The hand closes on the pan, the can or the toast in only 4 of 30 episodes.

\subsection{Real-World Experiments}
\label{sec:real_world}

\ourMethod carries over to a real robot hand. We train it on teleoperated play data~\citep{lynch2020learning}, about 50 demonstrations per skill, and use GPT-6 Astra as the behavior policy, as in \Cref{sec:agentic_policies}.
\\
\textbf{Tasks.}
We evaluate five tasks of increasing difficulty (\Cref{fig:real_tasks}).
\textbf{Pick pen} and \textbf{Pick-and-Place duck} involve objects that are easy to grasp.
\textbf{Pick-and-Place tomatoes} needs a precise pinch.
\textbf{Spray olive oil} and \textbf{Move mozzarella with tongs} are tool use tasks, where the hand must first grasp the tool precisely and then use it.
We also chain the tongs, tomato and spray tasks into one long-horizon episode that makes a salad (\Cref{fig:real_tasks}f), which we show qualitatively.
\\
\textbf{Results.}
We run 5 trials per task with GPT-6 Astra executed directly and 5 through \ourMethod.
Through \ourMethod, GPT-6 Astra succeeds in at least four of five trials on every task, and in all five on the duck and the tomatoes (\Cref{fig:real_results}). Executed directly, it solves the duck and the pen in most trials, since both tolerate an imprecise grasp. It succeeds once in five trials on the tomatoes and never on the two tool use tasks.

\section{Conclusion and Limitations}

We presented \ourMethod, a hand-object interaction controller that translates coarse behavioral commands into precise dexterous actions.
By conditioning on a semantic skill token, a coarse motion trajectory, wrist-depth imagery, and finger proprioception, \ourMethod corrects the last-mile errors that cause manipulation to fail near contact.
In simulation, \ourMethod maintains high task success under centimeter-scale position errors and large approach-direction deviations where direct execution drops to near zero.
Integrated with GPT-6 Astra and Claude Opus 5.5, \ourMethod enables agentic models to solve contact-rich tasks such as spraying, toasting, and tool use that they rarely solve alone, in simulation and, with GPT-6 Astra, on a real robot hand.
To the best of our knowledge, this is the first demonstration of agentic models performing dexterous manipulation in the real world.

Two limitations point to future work.
First, agentic models are bottlenecked by inference latency. Each decision step takes on the order of 15 seconds, making closed-loop reactive control impractical for time-sensitive tasks. Faster inference and tighter agent-controller coupling are needed.
Second, \ourMethod demonstrates spatial generalization (e.g., learning to spray circles and generalizing to arbitrary spray patterns) but has not been trained at scale across object categories. The objects seen at test time are the same as those seen during training, so \ourMethod cannot yet pick \emph{any} object or use \emph{any} spray bottle.
Scaling the training data and building generalist \glspl{hoic} that transfer across diverse objects is a key direction ahead.

\bibliographystyle{plainnat}
\bibliography{references}

@article{qian2024pianomime,
  title={Pianomime: Learning a generalist, dexterous piano player from internet demonstrations},
  author={Qian, Cheng and Urain, Julen and Zakka, Kevin and Peters, Jan},
  journal={arXiv preprint arXiv:2407.18178},
  year={2024}
}

@misc{pink,
  title = {{Pink: Python inverse kinematics based on Pinocchio}},
  author = {Caron, Stéphane and De Mont-Marin, Yann and Budhiraja, Rohan and Bang, Seung Hyeon and Domrachev, Ivan and Nedelchev, Simeon and Du, Peter and Escande, Adrien and Vaillant, Joris and Wingo, Bruce and Patapati, Santosh and San José Pro, Daniel and Marticorena Vidal, Nicolas Guillermo},
  license = {Apache-2.0},
  url = {https://github.com/pink-kinematics/pink},
  version = {4.4.0},
  year = {2026}
}

@inproceedings{dasari2024ditpi,
  title={The ingredients for robotic diffusion transformers},
  author={Dasari, Sudeep and Mees, Oier and Zhao, Sebastian and Srirama, Mohan Kumar and Levine, Sergey},
  booktitle={2025 IEEE International Conference on Robotics and Automation (ICRA)},
  pages={15617--15625},
  year={2025},
  organization={IEEE}
}

@article{couairon2022diffedit,
  title={Diffedit: Diffusion-based semantic image editing with mask guidance},
  author={Couairon, Guillaume and Verbeek, Jakob and Schwenk, Holger and Cord, Matthieu},
  journal={arXiv preprint arXiv:2210.11427},
  year={2022}
}

@article{meng2021sdedit,
  title={Sdedit: Guided image synthesis and editing with stochastic differential equations},
  author={Meng, Chenlin and He, Yutong and Song, Yang and Song, Jiaming and Wu, Jiajun and Zhu, Jun-Yan and Ermon, Stefano},
  journal={arXiv preprint arXiv:2108.01073},
  year={2021}
}

@inproceedings{he2024learning,
  title={Learning human-to-humanoid real-time whole-body teleoperation},
  author={He, Tairan and Luo, Zhengyi and Xiao, Wenli and Zhang, Chong and Kitani, Kris and Liu, Changliu and Shi, Guanya},
  booktitle={2024 IEEE/RSJ International Conference on Intelligent Robots and Systems (IROS)},
  pages={8944--8951},
  year={2024},
  organization={IEEE}
}

@article{yin2025dexteritygen,
  title={Dexteritygen: Foundation controller for unprecedented dexterity},
  author={Yin, Zhao-Heng and Wang, Changhao and Pineda, Luis and Hogan, Francois and Bodduluri, Krishna and Sharma, Akash and Lancaster, Patrick and Prasad, Ishita and Kalakrishnan, Mrinal and Malik, Jitendra and others},
  journal={arXiv preprint arXiv:2502.04307},
  year={2025}
}

@inproceedings{singh2024dextrah,
  title={Dextrah-rgb: Visuomotor policies to grasp anything with dexterous hands},
  author={Singh, Ritvik and Allshire, Arthur and Handa, Ankur and Ratliff, Nathan and Van Wyk, Karl},
  booktitle={2026 IEEE International Conference on Robotics and Automation (ICRA)},
  pages={5086--5093},
  year={2026},
  organization={IEEE}
}

@article{agarwal2023dexterous,
  title={Dexterous functional grasping},
  author={Agarwal, Ananye and Uppal, Shagun and Shaw, Kenneth and Pathak, Deepak},
  journal={arXiv preprint arXiv:2312.02975},
  year={2023}
}

@article{lum2024dextrah,
  title={Dextrah-g: Pixels-to-action dexterous arm-hand grasping with geometric fabrics},
  author={Lum, Tyler Ga Wei and Matak, Martin and Makoviychuk, Viktor and Handa, Ankur and Allshire, Arthur and Hermans, Tucker and Ratliff, Nathan D and Van Wyk, Karl},
  journal={arXiv preprint arXiv:2407.02274},
  year={2024}
}

@article{huang2024fungraspfunctionalgraspingdiverse,
  title={Fungrasp: Functional grasping for diverse dexterous hands},
  author={Huang, Linyi and Zhang, Hui and Wu, Zijian and Christen, Sammy and Song, Jie},
  journal={IEEE Robotics and Automation Letters},
  volume={10},
  number={6},
  pages={6175--6182},
  year={2025},
  publisher={IEEE}
}

@inproceedings{bharadhwaj2024track2act,
  title={Track2act: Predicting point tracks from internet videos enables generalizable robot manipulation},
  author={Bharadhwaj, Homanga and Mottaghi, Roozbeh and Gupta, Abhinav and Tulsiani, Shubham},
  booktitle={European Conference on Computer Vision},
  pages={306--324},
  year={2024},
  organization={Springer}
}

@article{chi2025diffusion,
  title={Diffusion policy: Visuomotor policy learning via action diffusion},
  author={Chi, Cheng and Xu, Zhenjia and Feng, Siyuan and Cousineau, Eric and Du, Yilun and Burchfiel, Benjamin and Tedrake, Russ and Song, Shuran},
  journal={The International Journal of Robotics Research},
  volume={44},
  number={10-11},
  pages={1684--1704},
  year={2025},
  publisher={Sage Publications, Sage UK: London, England}
}

@article{wococo,
  title={Wococo: Learning whole-body humanoid control with sequential contacts},
  author={Zhang, Chong and Xiao, Wenli and He, Tairan and Shi, Guanya},
  journal={arXiv preprint arXiv:2406.06005},
  year={2024}
}

@article{weng2024dexdiffuser,
  title={Dexdiffuser: Generating dexterous grasps with diffusion models},
  author={Weng, Zehang and Lu, Haofei and Kragic, Danica and Lundell, Jens},
  journal={IEEE Robotics and Automation Letters},
  volume={9},
  number={12},
  pages={11834--11840},
  year={2024},
  publisher={IEEE}
}

@article{xu2024flow,
  title={Flow as the cross-domain manipulation interface},
  author={Xu, Mengda and Xu, Zhenjia and Xu, Yinghao and Chi, Cheng and Wetzstein, Gordon and Veloso, Manuela and Song, Shuran},
  journal={arXiv preprint arXiv:2407.15208},
  year={2024}
}

@inproceedings{lynch2020learning,
  title={Learning latent plans from play},
  author={Lynch, Corey and Khansari, Mohi and Xiao, Ted and Kumar, Vikash and Tompson, Jonathan and Levine, Sergey and Sermanet, Pierre},
  booktitle={Conference on robot learning},
  pages={1113--1132},
  year={2020},
  organization={Pmlr}
}

@article{dexsuite,
  title={DexSuite: An All-in-One Framework for Dexterous Manipulation Across Arms, Hands, and Tasks},
  author={Houssaini, Anas and He, Hugo and Shi, Junming and Abyaneh, Amin and Chahine, Ricardo and Wen, Shuo and Danesh, Mohamad H and Fr{\'e}chette, Charles-{\'E}tienne and Guzman, Mariana Sosa and Desrochers, Mathias and others},
  journal={arXiv preprint arXiv:2610.37799},
  year={2026}
}

@inproceedings{fu2023deep,
  title={Deep whole-body control: Learning a unified policy for manipulation and locomotion},
  author={Fu, Zipeng and Cheng, Xuxin and Pathak, Deepak},
  booktitle={Conference on Robot Learning},
  pages={138--149},
  year={2023},
  organization={PMLR}
}

@article{silver2018residual,
  title={Residual policy learning},
  author={Silver, Tom and Allen, Kelsey and Tenenbaum, Josh and Kaelbling, Leslie},
  journal={arXiv preprint arXiv:1812.06298},
  year={2018}
}

@inproceedings{johannink2019residual,
  title={Residual reinforcement learning for robot control},
  author={Johannink, Tobias and Bahl, Shikhar and Nair, Ashvin and Luo, Jianlan and Kumar, Avinash and Loskyll, Matthias and Ojea, Juan Aparicio and Solowjow, Eugen and Levine, Sergey},
  booktitle={2019 international conference on robotics and automation (ICRA)},
  pages={6023--6029},
  year={2019},
  organization={IEEE}
}

@inproceedings{peebles2023scalable,
  title={Scalable diffusion models with transformers},
  author={Peebles, William and Xie, Saining},
  booktitle={2023 IEEE/CVF International Conference on Computer Vision (ICCV)},
  pages={4172--4182},
  year={2023},
  organization={IEEE}
}

@article{brohan2023rt2,
  title={Rt-2: Vision-language-action models transfer web knowledge to robotic control},
  author={Brohan, Anthony and Brown, Noah and Carbajal, Justice and Chebotar, Yevgen and Chen, Xi and Choromanski, Krzysztof and Ding, Tianli and Driess, Danny and Dubey, Avinava and Finn, Chelsea and others},
  journal={arXiv preprint arXiv:2307.15818},
  year={2023}
}

@article{kim2024openvla,
  title={Openvla: An open-source vision-language-action model},
  author={Kim, Moo Jin and Pertsch, Karl and Karamcheti, Siddharth and Xiao, Ted and Balakrishna, Ashwin and Nair, Suraj and Rafailov, Rafael and Foster, Ethan and Lam, Grace and Sanketi, Pannag and others},
  journal={arXiv preprint arXiv:2406.09246},
  year={2024}
}

@article{black2024pi0,
  title={$\pi_0$: A Vision-Language-Action Flow Model for General Robot Control},
  author={Black, Kevin and Brown, Noah and Driess, Danny and Esmail, Adnan and Equi, Michael and Finn, Chelsea and Fusai, Niccolo and Groom, Lachy and Hausman, Karol and Ichter, Brian and others},
  journal={arXiv preprint arXiv:2410.24164},
  year={2024}
}

@article{geminirobotics2025,
  title={Gemini robotics: Bringing ai into the physical world},
  author={Team, Gemini Robotics and Abeyruwan, Saminda and Ainslie, Joshua and Alayrac, Jean-Baptiste and Arenas, Montserrat Gonzalez and Armstrong, Travis and Balakrishna, Ashwin and Baruch, Robert and Bauza, Maria and Blokzijl, Michiel and others},
  journal={arXiv preprint arXiv:2503.20020},
  year={2025}
}

@article{ahn2022saycan,
  title={Do as i can, not as i say: Grounding language in robotic affordances},
  author={Ahn, Michael and Brohan, Anthony and Brown, Noah and Chebotar, Yevgen and Cortes, Omar and David, Byron and Finn, Chelsea and Fu, Chuyuan and Gopalakrishnan, Keerthana and Hausman, Karol and others},
  journal={arXiv preprint arXiv:2204.01691},
  year={2022}
}

@inproceedings{liang2022codepolicies,
  title={Code as policies: Language model programs for embodied control},
  author={Liang, Jacky and Huang, Wenlong and Xia, Fei and Xu, Peng and Hausman, Karol and Ichter, Brian and Florence, Pete and Zeng, Andy},
  booktitle={2023 IEEE International conference on robotics and automation (ICRA)},
  pages={9493--9500},
  year={2023},
  organization={IEEE}
}

@article{huang2022innermonologue,
  title={Inner monologue: Embodied reasoning through planning with language models},
  author={Huang, Wenlong and Xia, Fei and Xiao, Ted and Chan, Harris and Liang, Jacky and Florence, Pete and Zeng, Andy and Tompson, Jonathan and Mordatch, Igor and Chebotar, Yevgen and others},
  journal={arXiv preprint arXiv:2207.05608},
  year={2022}
}

@article{huang2023voxposer,
  title={Voxposer: Composable 3d value maps for robotic manipulation with language models},
  author={Huang, Wenlong and Wang, Chen and Zhang, Ruohan and Li, Yunzhu and Wu, Jiajun and Fei-Fei, Li},
  journal={arXiv preprint arXiv:2307.05973},
  year={2023}
}

@article{zhang2026gpt6astra,
  title={An unexpected robot policy: Early evaluations of GPT-6 Astra on RoboDojo and beyond},
  author={Zhang, Wenbo and Wang, Kaixuan and Ouyang, Yutao and Huang, Xiaoyu and Li, Liyang and Su, Kailun and Jin, Weiyang and Chai, Wenhao and Liang, Haotian and Dou, Zhiyang and others},
  journal={arXiv preprint arXiv:2609.24170},
  year={2026}
}

@article{halo2026,
  title={Perceptive Humanoid Motion from Simulation},
  author={Hu, Siyuan and Abuduweili, Abulikemu and Meng, Yue and Liu, Shipeng and Leboutet, Quentin and Chen, Chen and M{\"u}ller, Matthias and Koltun, Vladlen},
  journal={arXiv preprint arXiv:2611.06908},
  year={2026}
}

@article{gupta2026lucid,
  title={LUCID: Learning Embodiment-Agnostic Intent Models from Unstructured Human Videos for Scalable Dexterous Robot Skill Acquisition},
  author={Gupta, Harsh and Shi, Guanya and Yuan, Wenzhen},
  journal={arXiv preprint arXiv:2606.11628},
  year={2026}
}

@article{yoneda2023noise,
  title={To the noise and back: Diffusion for shared autonomy},
  author={Yoneda, Takuma and Sun, Luzhe and Yang, Ge and Stadie, Bradly and Walter, Matthew},
  journal={arXiv preprint arXiv:2302.12244},
  year={2023}
}

@article{chen2026robodojo,
  title={RoboDojo: A unified sim-and-real benchmark for comprehensive evaluation of generalist robot manipulation policies},
  author={Chen, Tianxing and Chen, Yue and Li, Zixuan and Tang, Junyuan and Su, Kailun and Lu, Haoran and Wan, Weijie and Chen, Baijun and Liu, Songling and Yan, Haowen and others},
  journal={arXiv preprint arXiv:2607.04434},
  year={2026}
}

@article{intelligence2025pi05,
  title={$\pi_{0.5}$: a Vision-Language-Action Model with Open-World Generalization},
  author={Intelligence, Physical and Black, Kevin and Brown, Noah and Darpinian, James and Dhabalia, Karan and Driess, Danny and Esmail, Adnan and Equi, Michael and Finn, Chelsea and Fusai, Niccolo and others},
  journal={arXiv preprint arXiv:2504.16054},
  year={2025}
}

@article{si2026pyrualean,
  title={Fewer tokens, better action: GPT-6 Astra robot agents with 14\% higher success rate but 65\% fewer tokens},
  author={Si, Ruiyang and Bi, Jianxin and Yang, Shunyu and Ni, Rui and Huang, Wenbo and Wang, Qiang and Jiang, Shulong and Zhang, Yintianrun and Ruan, Feixiang and Zhang, Kewei and others},
  journal={arXiv preprint arXiv:2610.01939},
  year={2026}
}

@article{gao2026hexashell,
  title={HexaShell: A Thin Harness for Exploring the Limits of VLM Agents as Robot Coding Agents},
  author={Gao, Hongcheng and Zhou, Alex and Zhu, Jay and Zheng, Zelin and Ge, Shijia and Li, Xiping and Wu, Di and Wang, Yazhe and Zeng, Jianshu and Shangguan, Xuan and others},
  journal={arXiv preprint arXiv:2610.23479},
  year={2026}
}

@article{li2026embodiedevo,
  title={EmbodiedEvo: Self-evolving coding agent for physical intelligence},
  author={Li, Wenyi and Chi, Yizhe and Zheng, Youjie and Liu, Chunyu and Wang, Xiaoqiu and Gao, Mingju and Jiang, Zheng and Hong, Deyao and Xiao, Calvin and Yang, Kaisen and others},
  journal={arXiv preprint arXiv:2610.18867},
  year={2026}
}

@inproceedings{park2026demodiffusion,
  title={Demodiffusion: One-shot human imitation using pre-trained diffusion policy},
  author={Park, Sungjae and Bharadhwaj, Homanga and Tulsiani, Shubham},
  booktitle={2026 IEEE International Conference on Robotics and Automation (ICRA)},
  pages={16884--16891},
  year={2026},
  organization={IEEE}
}

@misc{openai2026gpt6astra,
  title={{GPT-6 Astra System Card}},
  author={{OpenAI}},
  year={2026},
  url={https://deploymentsafety.openai.com/gpt-6-astra}
}

@misc{anthropic2026opus55,
  title={{Claude Opus 5.5 System Card}},
  author={{Anthropic}},
  year={2026},
  url={https://www.anthropic.com/claude-opus-5-5-system-card}
}

\onecolumn
\section*{Appendix}
\newcolumntype{Y}{>{\raggedright\arraybackslash}X}
\setcounter{section}{0}
\renewcommand{\thesection}{\Alph{section}}
\renewcommand{\thesubsection}{\thesection.\arabic{subsection}}
\renewcommand{\thesubsectiondis}{\thesubsection.}
\renewcommand{\theHsection}{app.\Alph{section}}

\section{Implementation Details}
\label{app:implementation}

\subsection{Architecture}
\begin{center}
\small
\begin{tabularx}{\textwidth}{lX}
\hline
Backbone & Encoder-decoder diffusion transformer, width 512, 8 heads \\
Encoder & 6 layers over 3 tokens (wrist depth, skill, proprioception) \\
Decoder & 6 DiT blocks over 40 action tokens \\
Conditioning & AdaLN on the flow time and the mean encoder token \\
Wrist depth & $224 \times 224$, clipped at 0.30~m, ResNet-18 from scratch \\
Proprioception & Finger joint angles (16 Allegro, 22 Sharpa) \\
Skill token & Learned embedding per skill, plus a no-skill token \\
Action chunk & 40 steps at 20~Hz (2~s) of wrist position relative to the current pose, rotation vector and finger joint angles \\
Generative model & Rectified flow, 8 Euler steps at inference \\
\hline
\end{tabularx}
\captionof{table}{\ourMethod network.}
\label{tab:app_architecture}
\end{center}

\subsection{Training Data}
\begin{center}
\small
\begin{tabularx}{\textwidth}{lllX}
\hline
Task & Hand & Episodes & Skills \\
\hline
\multicolumn{4}{l}{\textit{Simulation, scripted controller}} \\
Duck & Allegro & 100 & reach, pinch, carry, release \\
Pan & Allegro & 100 & reach, grasp, carry, release \\
Spray & Sharpa & 100 & reach, grasp, carry, press \\
Toaster & Sharpa & 100 & reach, grasp, carry, press, release, insert \\
\hline
\multicolumn{4}{l}{\textit{Real world, teleoperation}} \\
Pen & Sharpa & 50 & reach, grasp, carry, release \\
Duck & Sharpa & 50 & reach, grasp, carry, release \\
Tomatoes & Sharpa & 50 & reach, pick, carry, place \\
Spray & Sharpa & 50 & reach, grab, carry, press, release \\
Tongs & Sharpa & 50 & reach, grab, carry, close, open, release \\
\hline
\end{tabularx}
\captionof{table}{Training data.}
\label{tab:app_data}
\end{center}

\subsection{Training}
\begin{center}
\small
\begin{tabularx}{\textwidth}{lX}
\hline
Optimizer & AdamW, learning rate $10^{-4}$, weight decay $10^{-6}$ \\
Batch size & 96 \\
Loss & Flow velocity error, wrist weighted 4$\times$ \\
Skill relabeling & Grasp chunks become reach with probability 0.25 \\
Hardware & One NVIDIA L40S \\
\hline
\end{tabularx}
\captionof{table}{Training.}
\label{tab:app_training}
\end{center}

\subsection{Inference}
\begin{center}
\small
\begin{tabularx}{\textwidth}{lXX}
\hline
Initial sample $\mA_0$ & Training & Inference \\
\hline
Wrist, free and in-hand & Target wrist path & Behavior trajectory \\
Wrist, approach & Gaussian process (GP) sample, length scale 0.1 to 2~s & Behavior trajectory \\
Fingers & $\gN(\mathbf{0}, \mI)$ & $\gN(\mathbf{0}, \mI)$ \\
\hline
\end{tabularx}
\captionof{table}{Initial sample of the flow.}
\label{tab:app_inference}
\end{center}

\subsection{Synthetic Augmentation}
\begin{center}
\small
\begin{tabularx}{\textwidth}{lX}
\hline
Applied to & Free and in-hand chunks only \\
Paths per chunk & 1024 \\
Direction & One axis (half) or uniform, length up to the data maximum \\
Wiggle & GP, std up to 4~cm, length scale 0.3~s \\
Rotation & Ramp to $\theta \sim \mathcal{U}(0.05, 1)$ of the data maximum, wobble up to 0.05~rad \\
Fingers & Recorded \\
\hline
\end{tabularx}
\captionof{table}{Synthetic wrist paths.}
\label{tab:app_synthetic}
\end{center}

\section{Agentic Interface}
\label{app:agentic}

\subsection{Prompts}
Every agentic run starts from one base prompt. It describes the robot and its hand, the head and wrist cameras, the robot's frame, the answer format and the robot's reply after each step. With \ourMethod, the agent picks one of the task's skills. Alone, it sets the joint angles after 1 and 2~s instead. Each episode then opens with a task sentence and advice.

\subsection{Observations}
\begin{figure}[ht]
\centering
\includegraphics[width=\textwidth]{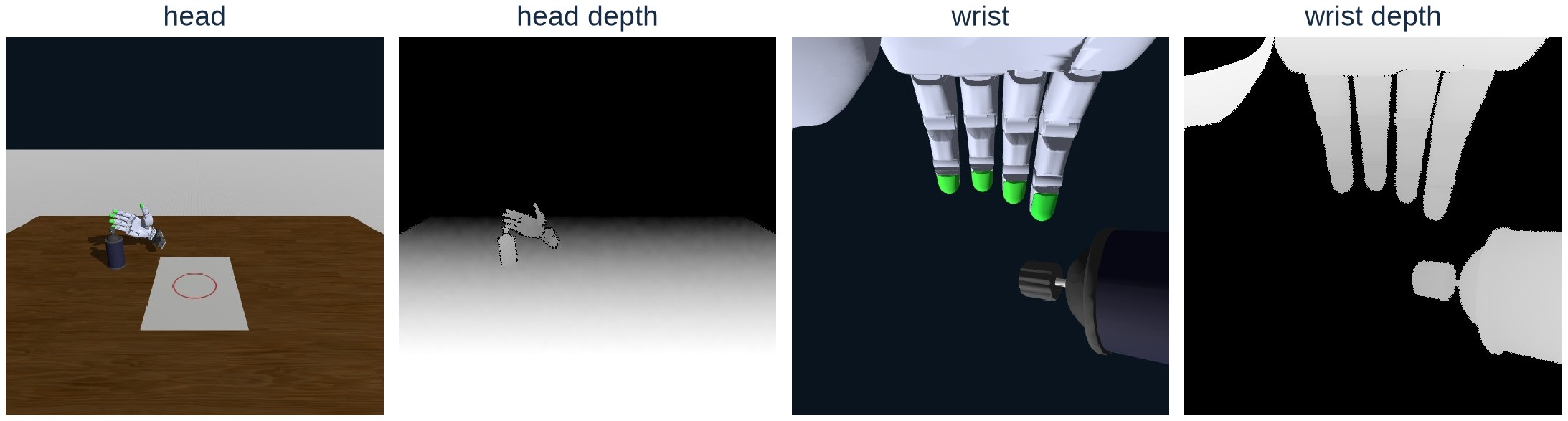}
\footnotesize
\begin{tabularx}{\textwidth}{>{\raggedright\arraybackslash}X>{\raggedright\arraybackslash}X}
\hline
\textbf{What the agent receives} & \textbf{What it answers} \\
\hline
Robot reply to turn 4: The robot ran the first 10 steps of your path. Relative to this turn's M0 the palm point ended at (-1.6, -2.1, -4.6) cm, turned (9, -10, -4) deg [\ldots] 0.4 cm and 3 deg apart [\ldots]\par\smallskip This is turn 5 of at most 60.\par\smallskip Robot now: fingers point (+0.35, -0.74, +0.58), palm faces (+0.91, +0.42, -0.02), thumb side (-0.22, +0.53, +0.82).\par\smallskip The current head and wrist images follow. & say ``I will start grasping now and advance slowly toward the can’s back without changing the hand angle.'' \newline
skill grasp \newline
mid xyz (4, -2.4, 0), rot (0, 0, 0) \newline
end xyz (8, -4.8, 0), rot (0, 0, 0) \newline
done false \newline
reason ``First closing-approach turn from the clear start pose.'' \\
\hline
\end{tabularx}
\caption{Turn 5 of GPT-6 Astra with \ourMethod on the spray task, zero-shot, with the message shortened.}
\label{fig:app_observation}
\end{figure}

At every step the agent receives the robot's reply to its last turn and the finger joint angles, with a colour and a depth image from the head and wrist cameras (\Cref{fig:app_observation}). It receives the same with and without \ourMethod.

\subsection{From Waypoints to a Trajectory}
The plan runs over 40 steps at 20~Hz. Its position follows a quadratic curve through the current palm pose $M_0$, mid and end, and its rotation turns by slerp from $M_0$ to mid and from mid to end. Each step moves the palm at most 1.5~cm along each axis and turns it at most 4.5$^\circ$ about each axis (30~cm/s and 90$^\circ$/s). The robot executes the first 10 steps (0.5~s). With \ourMethod, this plan is the wrist of the initial sample and the robot executes the first 10 steps of \ourMethod's output.

\subsection{Protocols}
\begin{center}
\small
\begin{tabularx}{\textwidth}{lYY}
\hline
 & Zero-shot execution & Iterative in-context learning \\
\hline
Context & Empty every episode & One conversation, up to 20 episodes \\
Episodes & 10 per task & Until learned, at most 20 \\
Feedback & None & The success check and final images after each episode, then the agent's notes \\
Learned & & Three successes in a row \\
Runs & One per condition & Three per condition \\
Episode end & \multicolumn{2}{l}{Agent says done or step limit (pan 35, duck 60, spray 60, toaster 80)} \\
Success & \multicolumn{2}{l}{Environment check at any step} \\
\hline
\end{tabularx}
\captionof{table}{Protocols.}
\label{tab:app_protocols}
\end{center}

\subsection{Models}
\begin{center}
\small
\begin{tabularx}{\textwidth}{llY}
\hline
Agent & Model & Settings \\
\hline
GPT-6 Astra & \texttt{gpt-6-astra} & Reasoning effort high \\
Claude Opus 5.5 & \texttt{claude-opus-5-5} & Adaptive thinking, effort xhigh \\
\hline
\end{tabularx}
\captionof{table}{Models.}
\label{tab:app_models}
\end{center}

\section{Simulation Tasks}
\label{app:sim_tasks}
All tasks run in DexSuite~\citep{dexsuite} at 20~Hz control with a 0.01~s physics step, with the hand floating on a 6-DoF wrist. The agent sees $512 \times 512$ colour and depth images from the head and wrist cameras.

\begin{center}
\footnotesize
\begin{tabularx}{\textwidth}{llYY}
\hline
Task & Hand & Objects & Success \\
\hline
Duck & Allegro (16) & Rubber duck, bowl & Duck still in the bowl, untouched \\
Pan & Allegro (16) & Pan, hot plate, mat & Pan on the mat \\
Spray & Sharpa (22) & Spray can, paper with a 15~cm circle & $\geq$60\% of the circle painted, $\leq$35\% outside \\
Toaster & Sharpa (22) & Toaster, toast & Toast in a slot, $\geq$3~cm deep, lever $\geq$80\% down \\
\hline
\end{tabularx}
\captionof{table}{Simulation tasks.}
\label{tab:app_sim_tasks}
\end{center}

\section{Noise Configuration}
\label{app:noise}

All errors corrupt the wrist trajectory of the behavior policy, only while the hand approaches the object. The behavior policy is a scripted controller that replans from the true state every 5 control steps.

\begin{table}[ht]
\centering
\begin{tabular}{lll}
\hline
Error & Description & Magnitudes \\
\hline
Position bias & constant shift in a random direction per episode & 2, 3, 5, 10 cm \\
Drift & Ornstein-Uhlenbeck offset, time constant 2 s & 2, 3, 5 cm (std per axis) \\
Approach direction & tilt of the approach path about the wrist & 5, 10, 20, 30, 45$^\circ$ \\
Jitter & white noise on every step of the trajectory & 0.2, 0.4, 0.6, 0.8 (std, normalized actions) \\
\hline
\end{tabular}
\caption{Error types used in \Cref{sec:imperfect_policies}.}
\label{tab:noise}
\end{table}

\paragraph{Protocol.} For every error type and magnitude, we run 10 episodes on seeds unseen during training, for the behavior policy executed directly and routed through \ourMethod (500k training steps, five training seeds). The behavior policy is evaluated on five such sets of 10 episodes. The curves of \Cref{fig:behavior_degradation} scale each error type to a relative severity between 0 (no error) and 1 (its largest magnitude), linearly interpolate the success of each type at the union of the measured severities, and average the four types.

\begin{table}[ht]
\centering
\small
\setlength{\tabcolsep}{2.8pt}
\begin{tabular}{lcccccccc}
\hline
 & \multicolumn{2}{c}{Pick-and-place duck} & \multicolumn{2}{c}{Pick-and-place pan} & \multicolumn{2}{c}{Spray graffiti} & \multicolumn{2}{c}{Toaster} \\
Error & Behavior policy & + \ourMethod & Behavior policy & + \ourMethod & Behavior policy & + \ourMethod & Behavior policy & + \ourMethod \\
\hline
No error & $100 \pm 0$ & $78 \pm 4$ & $100 \pm 0$ & $100 \pm 0$ & $100 \pm 0$ & $100 \pm 0$ & $100 \pm 0$ & $90 \pm 7$ \\
Position bias 2 cm & $64 \pm 19$ & $80 \pm 10$ & $98 \pm 4$ & $100 \pm 0$ & $46 \pm 22$ & $100 \pm 0$ & $30 \pm 7$ & $90 \pm 7$ \\
Position bias 3 cm & $38 \pm 18$ & $84 \pm 9$ & $88 \pm 11$ & $100 \pm 0$ & $32 \pm 8$ & $98 \pm 4$ & $2 \pm 4$ & $88 \pm 8$ \\
Position bias 5 cm & $12 \pm 8$ & $80 \pm 10$ & $50 \pm 12$ & $100 \pm 0$ & $18 \pm 4$ & $98 \pm 4$ & $0 \pm 0$ & $80 \pm 7$ \\
Position bias 10 cm & $4 \pm 5$ & $82 \pm 8$ & $26 \pm 11$ & $100 \pm 0$ & $4 \pm 5$ & $100 \pm 0$ & $2 \pm 4$ & $84 \pm 15$ \\
Drift 2 cm & $56 \pm 11$ & $82 \pm 11$ & $86 \pm 5$ & $100 \pm 0$ & $52 \pm 26$ & $100 \pm 0$ & $40 \pm 7$ & $90 \pm 17$ \\
Drift 3 cm & $30 \pm 10$ & $86 \pm 17$ & $72 \pm 8$ & $100 \pm 0$ & $16 \pm 13$ & $100 \pm 0$ & $20 \pm 7$ & $92 \pm 8$ \\
Drift 5 cm & $6 \pm 5$ & $82 \pm 13$ & $48 \pm 13$ & $100 \pm 0$ & $8 \pm 8$ & $100 \pm 0$ & $12 \pm 11$ & $88 \pm 8$ \\
Approach direction 5$^\circ$ & $98 \pm 4$ & $82 \pm 8$ & $100 \pm 0$ & $100 \pm 0$ & $82 \pm 16$ & $100 \pm 0$ & $70 \pm 28$ & $76 \pm 11$ \\
Approach direction 10$^\circ$ & $72 \pm 8$ & $80 \pm 12$ & $98 \pm 4$ & $100 \pm 0$ & $32 \pm 16$ & $100 \pm 0$ & $52 \pm 27$ & $84 \pm 9$ \\
Approach direction 20$^\circ$ & $20 \pm 10$ & $88 \pm 8$ & $72 \pm 16$ & $100 \pm 0$ & $10 \pm 12$ & $98 \pm 4$ & $42 \pm 20$ & $76 \pm 9$ \\
Approach direction 30$^\circ$ & $10 \pm 7$ & $86 \pm 11$ & $50 \pm 7$ & $100 \pm 0$ & $0 \pm 0$ & $100 \pm 0$ & $28 \pm 19$ & $80 \pm 12$ \\
Approach direction 45$^\circ$ & $2 \pm 4$ & $82 \pm 8$ & $26 \pm 15$ & $100 \pm 0$ & $0 \pm 0$ & $100 \pm 0$ & $22 \pm 25$ & $82 \pm 8$ \\
Jitter 0.2 & $20 \pm 7$ & $90 \pm 12$ & $90 \pm 7$ & $100 \pm 0$ & $52 \pm 13$ & $100 \pm 0$ & $58 \pm 13$ & $90 \pm 7$ \\
Jitter 0.4 & $4 \pm 5$ & $86 \pm 11$ & $50 \pm 19$ & $100 \pm 0$ & $2 \pm 4$ & $100 \pm 0$ & $2 \pm 4$ & $92 \pm 8$ \\
Jitter 0.6 & $0 \pm 0$ & $80 \pm 16$ & $28 \pm 20$ & $100 \pm 0$ & $0 \pm 0$ & $100 \pm 0$ & $2 \pm 4$ & $90 \pm 7$ \\
Jitter 0.8 & $0 \pm 0$ & $76 \pm 15$ & $16 \pm 9$ & $100 \pm 0$ & $0 \pm 0$ & $100 \pm 0$ & $0 \pm 0$ & $94 \pm 9$ \\
\hline
\end{tabular}
\caption{Task success (\%) for every error type and magnitude, 10 episodes each. Mean and standard deviation over five training seeds for \ourMethod and over five sets of evaluation episodes for the behavior policy.}
\label{tab:noise_results}
\end{table}

\section{Real-World Setup}
\label{app:real}
\Cref{tab:app_real_hw} lists the hardware.

\begin{center}
\small
\begin{tabularx}{\textwidth}{lY}
\hline
Arm & FAR dexbot arm, 7 joints \\
Hand & Sharpa Wave, 22 DoF \\
Wrist camera & RealSense D405, depth from Fast Foundation Stereo \\
Head camera & ZED \\
Teleoperation & MANUS glove for the hand, Vive tracker for the wrist \\
Compute & One RTX 5090 (robot server, Fast Foundation Stereo and \ourMethod) \\
Rates & State 100~Hz, inverse kinematics 120~Hz, \ourMethod 20~Hz \\
\hline
\end{tabularx}
\captionof{table}{Real-world hardware.}
\label{tab:app_real_hw}
\end{center}

\end{document}